\documentclass[11pt]{article}
\usepackage{acl2015}
\usepackage{times}
\usepackage{latexsym}
\usepackage{amsmath}
\usepackage{multirow}
\usepackage{url}
\usepackage{float}

\usepackage{mathrsfs}
\usepackage{graphicx}
\usepackage{amssymb}
\usepackage{algorithm}
\usepackage{algorithmic}
\usepackage{geometry}
\usepackage{amsthm}
\usepackage{mathtools}
\usepackage{caption}
\usepackage{subcaption}

\renewcommand{\algorithmicrequire}{\textbf{Input:}}

\newcommand{\btheta}{\boldsymbol{\theta}}

\newcommand{\bphi}{\boldsymbol{\phi}}
\newcommand{\bPhi}{\boldsymbol{\Phi}}
\newcommand{\grad}{\nabla}
\newcommand{\mEp}{\mathbb{E}_{\btheta}} %model expectation
\newcommand{\mE}{\mathbb{E}_{\btheta_t}} %model expectation
\newcommand{\mEt}{\mathbb{E}_{\btheta_t,t}} %stochastic model expectation
\newcommand{\oE}{\widetilde{\mathbb{E}}} %empirical mean
\newcommand{\oEt}{\widetilde{\mathbb{E}_t}} %stochastic empirical mean

\title{Training Conditional Random Fields with Natural Gradient Descent}

\author{Yuan Cao\\
Center for Language \& Speech Processing\\
The Johns Hopkins University\\
Baltimore, MD, USA, 21218\\
{\tt yuan.cao@jhu.edu}}
\date{}

\begin{document}
\maketitle
\begin{abstract}
We propose a novel parameter estimation procedure that works efficiently for conditional random fields (CRF). This algorithm is an extension to the maximum likelihood estimation (MLE), using loss functions defined by Bregman divergences which measure the proximity between the model expectation and the empirical mean of the feature vectors. This leads to a flexible training framework from which multiple update strategies can be derived using natural gradient descent (NGD). We carefully choose the convex function inducing the Bregman divergence so that the types of updates are reduced, while making the optimization procedure more effective by transforming the gradients of the log-likelihood loss function. The derived algorithms are very simple and can be easily implemented on top of the existing stochastic gradient descent (SGD) optimization procedure, yet it is very effective as illustrated by experimental results.
\end{abstract}

\section{Introduction}
Graphical models are used extensively to solve NLP problems. One of the most popular models is the conditional random fields (CRF)~\cite{Lafferty:2001}, which has been successfully applied to tasks like shallow parsing~\cite{Sha:2003}, name entity recognition~\cite{McCallum:2003}, word segmentation~\cite{Peng:2003} etc., just to name a few.

While the modeling power demonstrated by CRF is critical for performance improvement, accurate parameter estimation of the model is equally important. As a general structured prediction problem, multiple training methods for CRF have been proposed corresponding to different choices of loss functions. For example, one of the common training approach is maximum likelihood estimation (MLE), whose loss function is the (negative) log-likelihood of the training data and can be optimized by algorithms like L-BFGS~\cite{Liu:1989}, stochastic gradient descent (SGD)~\cite{Bottou:1998}, stochastic meta-descent (SMD)~\cite{Schraudolph:1999}~\cite{Vishwanathan:2006} etc. If the structured hinge-loss is chosen instead, then the Passive-Aggressive (PA) algorithm~\cite{Crammer:06}, structured Perceptron~\cite{Collins:02} (corresponding to a hinge-loss with zero margin) etc. can be applied for learning.

In this paper, we propose a novel CRF training procedure. Our loss functions are defined by the Bregman divergence~\cite{Bregman:1967} between the model expectation and empirical mean of the feature vectors, and can be treated as a generalization of the log-likelihood loss. We then use natural gradient descent (NGD)~\cite{Amari:1998} to optimize the loss. Since for large-scale training, stochastic optimization is usually a better choice than batch optimization~\cite{Bottou:2008}, we focus on the stochastic version of the algorithms. The proposed framework is very flexible, allowing us to choose proper convex functions inducing the Bregman divergences that leads to better training performances.

 In Section~\ref{sec:background}, we briefly reviews some background materials that are relevant to further discussions; Section~\ref{sec:algorithm} gives a step-by-step introduction to the proposed algorithms; Experimental results are given in Section~\ref{sec:experiments}, followed by discussions and conclusions.

\section{Background}\label{sec:background}
\subsection{MLE for Graphical Models}\label{bg:mle}
Graphical models can be naturally viewed as exponential families~\cite{Wainwright:2008}. For example, for a data sample $(x,y)$ where $x$ is the input sequence and $y$ is the label sequence, the conditional distribution $p(y|x)$ modeled by CRF can be written as
\begin{eqnarray*}
p_{\btheta}(y|x) = \exp \{ \btheta\cdot\bPhi(x,y)-A_{\btheta} \} 
\end{eqnarray*}
where $\bPhi(x,y)=\sum\limits_{c\in\mathcal{C}} \bphi_c(x_c,y_c) \in \mathbb{R}^d$ is the feature vector (of dimension $d$) collected from all factors $\mathcal{C}$ of the graph, and $A_{\btheta}$ is the log-partition function.

MLE is commonly used to estimate the model parameters $\btheta$. The gradient of the log-likelihood of the training data is given by
\begin{eqnarray}\label{eq:grad}
\oE-\mathbb{E}_{\btheta}
\end{eqnarray}
where $\oE$ and $\mEp$ denotes the empirical mean of and model expectation of the feature vectors respectively. The moment-matching condition $\oE=\mathbb{E}_{\btheta^*}$ holds when the maximum likelihood solution $\btheta^*$ is found.
%define E, E_theta:
%$\oE\triangleq\frac{1}{N}\sum\limits_{i=1}^N \bPhi(x_i,y_i)$, $\mathbb{E}_{\btheta}\triangleq\frac{1}{N}\sum\limits_{i=1}^N\mathbb{E}_{p_{\btheta}(Y|x_i)}\left[ \bPhi(x_i,Y) \right]$

% namely we aim to find $\btheta$ that maximizes the (conditional) likelihood of the training data:
% \begin{eqnarray}\label{eq:likelihood}
% \frac{1}{N}\sum\limits_{i=1}^N \log p_{\btheta}(y_i|x_i)
% \end{eqnarray}

\subsection{Bregman Divergence}
The Bregman divergence is a general proximity measure between two points $\mathbf{p}, \mathbf{q}$ (can be scalars, vectors, or matrices). It is defined as
\begin{eqnarray*}
B_G(\mathbf{p}, \mathbf{q}) = G(\mathbf{p}) - G(\mathbf{q}) - \grad G(\mathbf{q})^T(\mathbf{p}-\mathbf{q})
\end{eqnarray*}
where $G$ is a differentiable convex function inducing the divergence. Note that Bregman divergence is in general asymmetric, namely $B_G(\mathbf{p}, \mathbf{q}) \neq B(\mathbf{q}, \mathbf{p})$.

By choosing $G$ properly, many interesting distances or divergences can be recovered. For example, choosing $G(\mathbf{u})=\frac{1}{2}\|\mathbf{u}\|^2$, $B_G(\mathbf{p}, \mathbf{q}) = \|\mathbf{p}-\mathbf{q}\|^2$ and the Euclidean distance is recovered; Choosing $G(\mathbf{u})=\sum\limits_i u_i \log u_i$, $B_G(\mathbf{p}, \mathbf{q}) = \sum\limits_i p_i \log\frac{p_i}{q_i}$ and the Kullback-Leibler divergence is recovered. A good review of the Bregman divergence and its properties can be found in~\cite{Banerjee:2005}.

%In this paper, we use the Bregman divergence to define loss functions that extends the log-likelihood loss. We assume $G$ to be at least twice-differentiable throughout the paper.

\subsection{Natural Gradient Descent}
NGD is derived from the study of information geometry~\cite{Amari:2000}, and is one of its most popular applications. The conventional gradient descent assumes the underlying parameter space to be Euclidean, nevertheless this is often not the case (say when the space is a statistical manifold). By contrast, NGD takes the geometry of the parameter space into consideration, giving an update strategy as follows:
\begin{eqnarray}\label{eq:ngd}
\btheta_{t+1} = \btheta_t - \lambda_t M_{\btheta_t}^{-1} \grad\ell(\btheta_t) 
\end{eqnarray}
where $\lambda_t$ is the learning rate and $M$ is the Riemannian metric of the parameter space manifold. When the parameter space is a statistical manifold, the common choice of $M$ is the Fisher information matrix, namely $M_{\btheta}^{i,j}=\mathbb{E}_p\left[ \frac{\partial\log p_{\btheta}(x)}{\partial\theta_i} \frac{\partial\log p_{\btheta}(x)}{\partial\theta_j}\right]$. It is shown in~\cite{Amari:1998} that NGD is asymptotically Fisher efficient and often converges faster than the normal gradient descent.

Despite the nice properties of NGD, it is in general difficult to implement due to the nontrivial computation of $M_{\btheta}^{-1}$. To handle this problem, researchers have proposed techniques to simplify, approximate, or bypass the computation of $M_{\btheta}^{-1}$, for example~\cite{LeRoux:2007}~\cite{Honkela:2008}~\cite{Pascanu:2013}.

%In this paper, we use NGD to optimize the loss functions defined by Bregman divergences. The computation of $M_{\btheta}^{-1}$ can be circumvented, thanks to the specific structure of the problem.

\section{Algorithm}\label{sec:algorithm}
\subsection{Loss Functions}\label{sec:loss}
The loss function defined for our training procedure is motivated by MLE. Since $\mEp=\oE$ needs to hold when the solution to MLE is found, the ``gap'' between $\mEp$ and $\oE$ (according to some measure) needs to be reduced as the training proceeds. We therefore define the Bregman divergence between $\mEp$ and $\oE$ as our loss function. Since the Bregman divergence is in general asymmetric, we consider four types of loss functions as follows (named $B_1$-$B_4$)\footnote{In the most general setting, we may use $B(a\mEp - b\oE, c\mEp - d\oE)\; s.t. \; a-b = c-d, \; a,b,c,d \in \mathbb{R}$ as the loss functions, which guarantees $\mEp=\oE$ holds at its minimum. However, this formulation brings too much design freedom which complicates the problem, since we are free to choose the parameters $a,b,c$ as well as the convex function $G$. Therefore, we narrow down our scope to the four special cases of the loss functions $B_1-B_4$ given above.}:
\begin{align*}
 & B_G\left(\gamma\mEp,\oE-\rho\mEp\right) && (B_1)\\
 & B_G\left(\oE-\rho \mEp, \gamma\mEp\right) && (B_2)
\end{align*}
\begin{align*}
 & B_G\left(\rho \oE, \mEp - \gamma\oE\right) && (B_3)\\
 & B_G\left(\mEp-\gamma \oE, \rho\oE\right) && (B_4)
\end{align*}
where $\gamma \in \mathbb{R}$ and $\rho \triangleq 1-\gamma$. It can be seen that whenever the loss functions are minimized at point $\btheta^*$ (Bregman divergence reaches zero), $\mathbb{E}_{\btheta^*}=\oE$ and $\btheta^*$ give the same solution as MLE.

We are free to choose the hyper-parameter $\gamma$ which is possibly correlated with the performance of the algorithm. However to simplify the problem setting, we will only focus on the cases where $\gamma=0$ or $1$. Although it seems we now have eight versions of loss functions, it will be seen shortly that by properly choosing the convex function $G$, many of them are redundant and we end up having only two update strategies.

\subsection{Applying Natural Gradient Descent}
The gradients of loss functions $B_1$-$B_4$ with respect to $\btheta$ are given in Table \ref{table:grads}.
\begin{table}[htbp]
  {\small
  \begin{tabular}{ c c c }
  Loss & Gradient wrt. $\btheta$ & $\gamma$ \\
  \hline
  \multirow{2}{1em}{$B_1$} & $\grad_{\btheta} \mEp \left[ \grad G(\mEp) - \grad G(\oE) \right]$ & $1$ \\
   & $\grad_{\btheta} \mEp \grad^2 G(\oE - \mEp) (\mEp - \oE) $ & $0$ \\
  \hline
  \multirow{2}{1em}{$B_2$} & $\grad_{\btheta} \mEp \grad^2 G(\mEp) (\mEp - \oE) $ & $1$\\
  & $\grad_{\btheta} \mEp \left[ \grad G(\mathbf{0}) - \grad G(\oE - \mEp) \right]$ & $0$ \\
  \hline
  $B_3$ & $\grad_{\btheta} \mEp \grad^2 G(\mEp-\gamma\oE) (\mEp - \oE) $ & $\{0,1\}$ \\
  \hline
  $B_4$ & $\grad_{\btheta} \mEp \left[ \grad G(\mEp-\gamma\oE ) - \grad G(\rho\oE) \right]$ & $\{0,1\}$ \\
  \hline
  \end{tabular} }
\caption{Gradients of loss functions $B_1$-$B_4$.} \label{table:grads}
\end{table}

It is in general difficult to compute the gradients of the loss functions, as all of them contain the term $\grad_{\btheta} \mEp$, whose computation is usually non-trivial. However, it turns out that the \emph{natural gradients} of the loss functions can be handled easily. To see this, note that for distributions in exponential family, $\grad_{\btheta} \mEp = \grad_{\btheta}^2 A_{\btheta_t} = M_{\btheta_t}$, which is the Fisher information matrix. Therefore the NGD update (Eq.~\ref{eq:ngd}) becomes
\begin{eqnarray}
\btheta_{t+1} = \btheta_t - \lambda_t \grad_{\btheta} \mE^{-1} \grad\ell(\btheta_t) \label{eq:ngd_exp}
\end{eqnarray}
Now if we plug, for example $\grad_{\btheta} B_1, \gamma=0$, into Eq.~\ref{eq:ngd_exp}, we have
\begin{eqnarray}
 \btheta_{t+1} &=& \btheta_t -\lambda_t \grad_{\btheta} \mE^{-1} \grad_{\btheta} B_2 \notag \\
 &=& \btheta_{t} - \lambda_t \grad^2 G(\oE - \mE) (\mE - \oE) \label{eq:update1}
\end{eqnarray}
Thus the step of computing the the Fisher information can be circumvented, making the optimization of our loss functions tractable\footnote{In ~\cite{Hoffman:2013}, a similar trick was also used. However their technique was developed from a specific variational inference problem setting, whereas the proposed approach derived from Bregman divergence is more general.}.

This trick applies to all gradients in Table \ref{table:grads}, yielding multiple update strategies. Note that $\grad_{\btheta}B_1, \gamma=1$ is equivalent to $\grad_{\btheta}B_4, \gamma=0$, and $\grad_{\btheta}B_2, \gamma=1$ is equivalent to $\grad_{\btheta}B_3, \gamma=0$. By applying Eq.~\ref{eq:ngd_exp} to all unique gradients, the following types of updates are derived (named $U_1$-$U_6$):
\begin{align*}
&\btheta_{t+1} = \btheta_t - \lambda_t \grad^2 G(\oE - \mE) (\mE - \oE) && (U_1) \\
&\btheta_{t+1} = \btheta_t - \lambda_t \left[ \grad G(\mathbf{0}) - \grad G(\oE - \mE) \right] && (U_2) \\
&\btheta_{t+1} = \btheta_t - \lambda_t \grad^2 G(\mE-\oE) (\mE - \oE) && (U_3) \\
&\btheta_{t+1} = \btheta_t - \lambda_t \grad^2 G(\mE) (\mE - \oE) && (U_4) \\
&\btheta_{t+1} = \btheta_t - \lambda_t \left[ \grad G(\mE-\oE ) - \grad G(\mathbf{0}) \right] && (U_5) \\
&\btheta_{t+1} = \btheta_t - \lambda_t \left[ \grad G(\mE) - \grad G(\oE) \right] && (U_6)
\end{align*}

\subsection{Reducing the Types of Updates}\label{sec:reduce}
Although the framework described so far is very flexible and multiple update strategies can be derived, it is not a good idea to try them all in turn for a given task. Therefore, it is necessary to reduce the types of updates and simplify the problem.

We first remove $U_4$ and $U_6$ from the list, since they can be recovered from $U_3$ and $U_5$ respectively by choosing $G'(\mathbf{u})=G(\mathbf{u}+\oE)$. To further reduce the update types, we impose the following constraints on the convex function $G$:
\begin{enumerate}
 \item $G$ is symmetric: $G(\mathbf{u}) = G(\mathbf{-u})$.
 \item $\grad G(\mathbf{u})$ is an element-wise function, namely $\grad G(\mathbf{u})_i = g_i(u_i), \forall i \in 1, \ldots, d$, where $g_i$ is a uni-variate scalar function.
\end{enumerate}

For example, $G(\mathbf{u})=\frac{1}{2}\|\mathbf{u}\|^2$ is a typical function satisfying the constraints, since $\frac{1}{2}\|\mathbf{u}\|^2=\frac{1}{2}\|-\mathbf{u}\|^2$, and $\grad G(\mathbf{u}) = [u_1,\ldots,u_d]^{T}$ where $g_i(u)=u$, $\forall i \in 1, \ldots, d$. It is also worth mentioning that by choosing $G(\mathbf{u})=\frac{1}{2}\|\mathbf{u}\|^2$, all updates $U_1$-$U_6$ become equivalent:
\[ \btheta_{t+1} = \btheta_t - \lambda_t \left(\mE - \oE\right) \]
which recovers the GD for the log-likelihood function.

When a twice-differentiable function $G$ satisfies the constraints 1 and 2, we have 
\begin{gather}
\grad G(\mathbf{u})=-\grad G(\mathbf{-u}) \label{eq:cond1} \\
\grad G(\mathbf{0}) = \mathbf{0} \label{eq:cond2} \\ 
\grad^2 G(\mathbf{u})=\grad^2 G(\mathbf{-u}) \label{eq:cond3}
\end{gather}
where Eq.~\ref{eq:cond3} holds since $\grad^2 G$ is a diagonal matrix. Given these conditions, we see immediately that $U_1$ is equivalent to $U_3$, and $U_2$ is equivalent to $U_5$. This way, the types of updates are eventually narrowed down to $U_1$ and $U_2$.

To see the relationship between $U_1$ and $U_2$, note that the Taylor expansion of $\grad G(\mathbf{0})$ at point $\oE - \mE$ is given by
{\small \begin{align*}
\grad G(\mathbf{0}) = & \grad G(\oE - \mE) + \grad^2 G(\oE - \mE) (\mE - \oE)\\
& + O(\|\mE - \oE\|^2)
\end{align*}}
Therefore $U_1$ and $U_2$ can be regarded as approximations to each other.

Since stochastic optimization is preferred for large-scale training, we replace $\oE$ and $\mE$ with its stochastic estimation $\oEt$ and $\mEt$, where $\oEt\triangleq\bPhi(x_t,y_t)$ and $\mEt\triangleq\mathbf{E}_{p_{\btheta_t}(Y|x_t)}\left[ \bPhi(x_t,Y) \right]$. Assuming $G$ satisfies the constraints, we re-write $U_1$ and $U_2$ as
\begin{align*}
&\btheta_{t+1} = \btheta_t - \lambda_t \grad^2 G(\mEt - \oEt) (\mE - \oEt) && (U_1^*) \\
&\btheta_{t+1} = \btheta_t - \lambda_t \grad G(\mEt - \oEt) && (U_2^*)
\end{align*}
which will be the focus for the rest of the paper.

\subsection{Choosing the Convex Function $G$}
We now proceed to choose the actual functional forms of $G$ aiming to make the training procedure more efficient.

A na{\"i}ve approach would be to choose $G$ from a parameterized convex function family (say the vector $p$-norm, where $p \geq 1$ is the hyper-parameter), and tune the hyper-parameter on a held-out set hoping to find a proper value that works best for the task at hand. However this approach is very inefficient, and we would like to choose $G$ in a more principled way.

Although we derived $U_1^*$ and $U_2^*$ from NGD, they can be treated as SGD of two surrogate loss functions $S_1(\btheta)$ and $S_2(\btheta)$ respectively (we do not even need to know what the actual functional forms of $S_1$, $S_2$ are), whose gradients $\grad_{\btheta} S_1 = \grad^2 G(\mathbb{E}_{\btheta} - \oE) (\mathbb{E}_{\btheta} - \oE)$ and $\grad_{\btheta} S_2 = \grad G(\mathbb{E}_{\btheta} - \oE)$ are transformations of the gradient of the log-likelihood (Eq.~\ref{eq:grad}). Since the performance of SGD is sensitive to the condition number of the objective function~\cite{Bottou:2008}, one heuristic for the selection of $G$ is to make the condition numbers of $S_1$ and $S_2$ smaller than that of the log-likelihood. However, this is hard to analyze since the condition number is in general difficult to compute. Alternatively, we may select a $G$ so that the second-order information of the log-likelihood can be (approximately) incorporated, as second-order stochastic optimization methods usually 
converge faster and are insensitive to the condition number of the objective function~\cite{Bottou:2008}. This is the guideline we follow in this section.

The first convex function we consider is as follows:
\begin{eqnarray*}
G_1(\mathbf{u})=\sum\limits_{i=1}^d\frac{u_i}{\sqrt{\epsilon}}\arctan(\frac{u_i}{\sqrt{\epsilon}}) - \frac{1}{2}\log\left(1+\frac{u_i^2}{\epsilon}\right)
\end{eqnarray*}
where $\epsilon>0$ is a small constant free to choose. It can be easily checked that $G_1$ satisfies the constraints imposed in Section~\ref{sec:reduce}. The gradients of $G_1$ are given by
\begin{gather*}
\grad G_1(\mathbf{u})= \frac{1}{\sqrt{\epsilon}}  \left[ \arctan(\frac{u_1}{\sqrt{\epsilon}}), \ldots, \arctan(\frac{u_d}{\sqrt{\epsilon}}) \right]^T\\
\grad^2 G_1(\mathbf{u}) = \text{diag}\left[ \frac{1}{u_1^2+\epsilon}, \ldots, \frac{1}{u_d^2+\epsilon} \right]
\end{gather*}

In this case, the $U_1^*$ update has the following form (named $U_1^*.G_1$):
{\small \begin{align*}
&\btheta_{t+1}^i = \btheta_t^i - \lambda_t \left[\left(\mEt^i - \oEt^i\right)^2+\epsilon\right]^{-1}\left(\mEt^i - \oEt^i\right)
\end{align*}}
where $i=1,\ldots,d$. This update can be treated as a stochastic second-order optimization procedure, as it scales the gradient Eq.~\ref{eq:grad} by the inverse of its variance in each dimension, and it reminds us of the online Newton step (ONS)~\cite{Hazan:2007} algorithm, which has a similar update step. However, in contrast to ONS where the full inverse covariance matrix is used, here we only use the diagonal of the covariance matrix to scale the gradient vector. Diagonal matrix approximation is often used in optimization algorithms incorporating second-order information (for example SGN-QN~\cite{Bordes:2009}, AdaGrad~\cite{Duchi:2011} etc.), as it can be computed orders-of-magnitude faster than using the full matrix ,without sacrificing much performance.

\begin{figure}
\centering
  \includegraphics[width=0.3\textwidth]{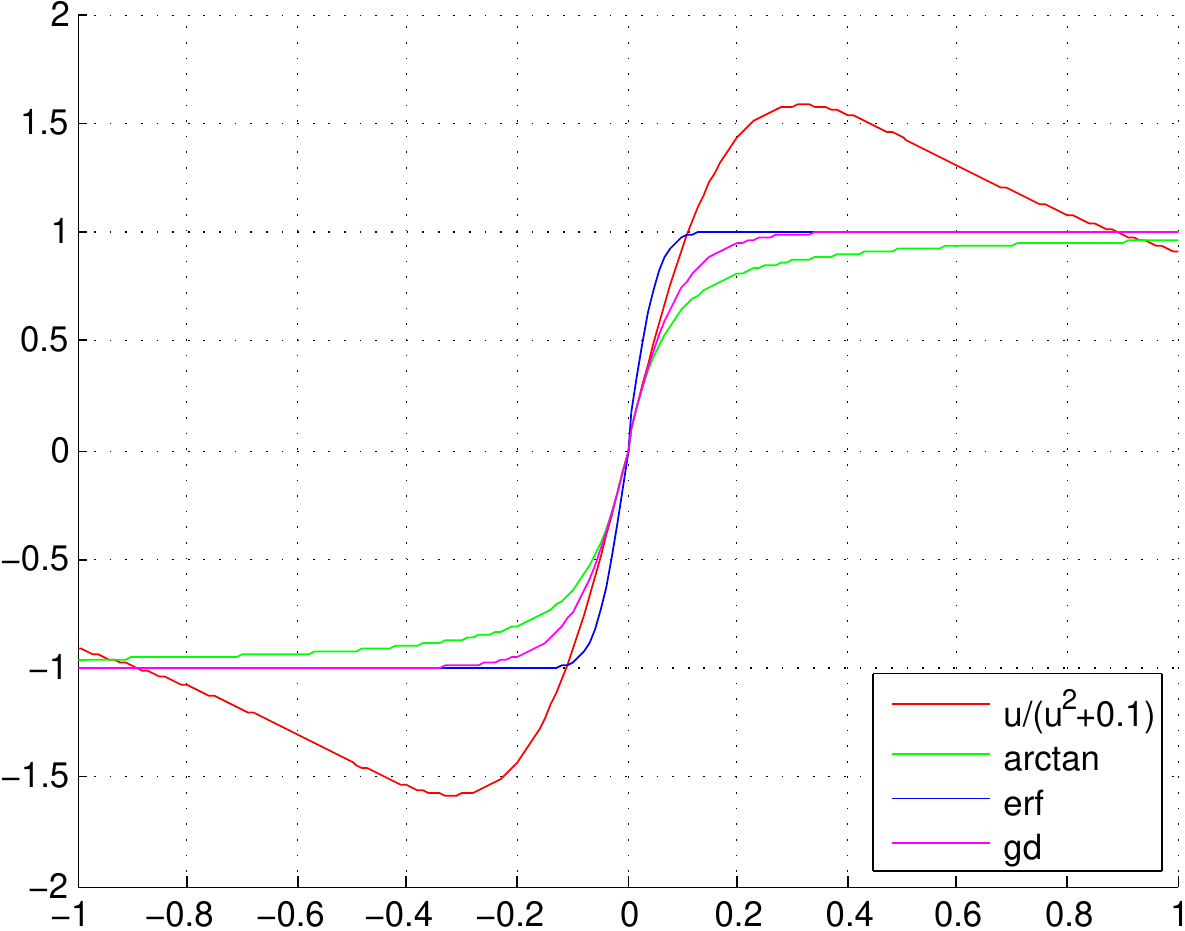}
  \caption{Comparison of functions $u/(u^2+0.1)$, \texttt{arctan}, \texttt{erf}, \texttt{gd}. The sigmoid functions \texttt{arctan}, \texttt{erf} and \texttt{gd} are normalized so that they have the same gradient as $u/(u^2+0.1)$ at $u=0$ and their values are bounded by -1, 1.} \label{fig:funcs}
\end{figure}

The $U_2^*$ update corresponding to the choice of $G_1$ has the following form (named $U_2^*.G_1$):
{\small \begin{align*}
&\btheta_{t+1}^i = \btheta_t^i - \lambda_t \arctan\left( \frac{\mEt^i - \oEt^i}{\sqrt{\epsilon}} \right)
\end{align*}}
Note that the constant $1/\sqrt{\epsilon}$ in $\grad G_1$ has been folded into the learning rate $\lambda_t$. Although not apparent at first sight, this update is in some sense similar to $U_1^*.G_1$. From $U_1^*.G_1$ we see that in each dimension, gradients $\mE^i-\oEt^i$ with smaller absolute values get boosted more dramatically than those with larger values (as long as $|\mE^i-\oEt^i|$ is not too close to zero). A similar property also holds for $U_2^*.G_1$, since \texttt{arctan} is a sigmoid function, and as long as we choose $\epsilon < 1$ so that
\begin{eqnarray*}
\frac{d}{du} \arctan\left(\frac{1}{\sqrt{\epsilon}}u\right) \rvert_{u=0} = \frac{1}{\sqrt{\epsilon}} > 1
\end{eqnarray*}
the magnitude of $\mE^i-\oEt^i$ with small absolute value will also get boosted. This is illustrated in Figure~\ref{fig:funcs} by comparing functions $\frac{u}{u^2+\epsilon}$ (where we select $\epsilon=0.1$) and $\arctan\left(\frac{1}{\sqrt{\epsilon}}u\right)$. Note that for many NLP tasks modeled by CRF, only indicator features are defined. Therefore the value of $\mE^i-\oEt^i$ is bounded by -1 and 1, and we only care about function values on this interval.
  
Since \texttt{arctan} belongs to the sigmoid function family, it is not the only candidate whose corresponding update mimics the behavior of $U_1^*.G_1$, while $G$ still satisfies the constraints given in Section ~\ref{sec:reduce}. We therefore consider two more convex functions $G_2$ and $G_3$, whose gradients are the \texttt{erf} and Gudermannian functions (\texttt{gd}) from the sigmoid family respectively:
\begin{gather*}
 \grad G_2(\alpha \mathbf{u})_i = \frac{2}{\pi}\int_{0}^{\alpha u_i} \exp\{-{x^2}\}dx\\
 \grad G_3(\beta \mathbf{u})_i = 2\arctan(\exp\{\beta u_i\}) - \frac{\pi}{2}
\end{gather*}
where $\alpha, \beta \in \mathbb{R}$ are hyper-parameters. In this case, we do not even know the functional form of the functions $G_2$, $G_3$, however it can be checked that both of them satisfy the constraints. The reason why we select these two functions from the sigmoid family is that when the gradients of \texttt{erf} and \texttt{d} are the same as that of \texttt{arctan} at point zero, both of them stay on top of \texttt{arctan}. Therefore, \texttt{erf} and \texttt{gd} are able to give stronger boosts to small gradients. This is also illustrated in Figure~\ref{fig:funcs}.

Applying $\grad G_2$ and $\grad G_3$ to $U_2^*$, we get two updates named $U_2^*.G_2$ and $U_2^*.G_3$ respectively. We do not consider further applying $\grad^2G_2$ and $\grad^3G_3$ to $U_1^*$, as the meaning of the updates becomes less clear.

%TODO: show that the objective is convex

\subsection{Adding Regularization Term}
So far we have not considered adding regularization term to the loss functions yet, which is crucial for better generalization. A common choice of the regularization term is the 2-norm (Euclidean metric) of the parameter vector $\frac{C}{2}\|\btheta\|^2$, where $C$ is a hyper-parameter specifying the strength of regularization. In our case, however, we derived our algorithms by applying NGD to the Bregman loss functions, which assumes the underlying parameter space to be non-Euclidean. Therefore, the regularization term we add has the form $\frac{C}{2}\btheta^T M_{\btheta} \btheta$, and the regularized loss to be minimized is
\begin{eqnarray}
\frac{C}{2}\btheta^T M_{\btheta} \btheta + B_i
\end{eqnarray}
where $B_i, i=\{1,2,3,4\}$ are the loss functions $B_1$-$B_4$.

However, the Riemannian metric $M_{\btheta}$ itself is a function of $\btheta$, and the gradient of the objective at time $t$ is difficult to compute. Instead, we use an approximation by keeping the Riemannian metric fixed at each time stamp, and the gradient is given by $M_{\btheta_t} \btheta_t + \grad_{\btheta_t} B_i$. Now if we apply NGD to this objective, the resulting updates will be no different than the SGD for L2-regularized surrogate loss functions $S_1$, $S_2$:
\begin{eqnarray*}
\btheta_{t+1} = (1-C\lambda_t)\left( \btheta_t - \frac{\lambda_t}{1-C\lambda_t} \grad_{\btheta}{S_{i,t}(\btheta_t)}\right)
\end{eqnarray*}
where $i=\{1,2\}$. It is well-known that SGD for L2-regularized loss functions has an equivalent but more efficient sparse update~\cite{Shalev-Schwartz:2007}~\cite{Bottou:2012}:
\begin{align*}
& \bar{\btheta}_t = \frac{\btheta_t}{z_t}\\
& \bar{\btheta}_{t+1} = \bar{\btheta}_t - \frac{\lambda_t}{(1-\lambda_t)z_t} \grad_{\btheta}{S_{i,t}(\btheta_t)}\\
& z_{t+1} = (1-C\lambda_t)z_t
\end{align*}
where $z_t$ is a scaling factor and $z_0=1$. We then modify $U_1^*$ and $U_2^*$ accordingly, simply by changing the step-size and maintaining a scaling factor.

As for the choice of learning rate $\lambda_t$, we follow the recommendation given by~\cite{Bottou:2012} and set $\lambda_t=\left[\hat{\lambda}(1+\hat{\lambda}Ct)\right]^{-1}$, where $\hat{\lambda}$ is calibrated from a small subset of the training data before the training starts.

The final versions of the algorithms developed so far are summarized in Figure~\ref{fig:alg}.
\begin{figure}[t!]
  \begin{algorithmic}
  \renewcommand{\algorithmicrequire}{\textbf{Initialization:}}
  \REQUIRE
  \STATE Choose hyper-parameters $C$, $\epsilon/\alpha/\beta$ (depending on which convex function to use: $G_1$, $G_2$, $G_3$).\\
   Set $z_0=1$, and calibrate $\hat{\lambda}$ on a small training subset.
  \renewcommand{\algorithmicrequire}{\textbf{Algorithm:}} 
  \REQUIRE
  \FOR{$e=1 \ldots \texttt{num epoch}$}
  \FOR{$t=1 \ldots T$}
  \STATE Receive training sample $(x_t,y_t)$
  \STATE $\lambda_t=\left[\hat{\lambda}(1+\hat{\lambda}Ct)\right]^{-1}$, $\btheta_t = z_t\bar{\btheta}_t$
  \STATE Depending on the update strategy selected, update the parameters:\\
  \STATE $\bar{\btheta}_{t+1} = \bar{\btheta}_t - \frac{\lambda_t}{(1-\lambda_t)z_t} \grad_{\btheta}{S_{t}(\btheta_t)}$\\
  \STATE where $\grad_{\btheta}{S_{t}(\btheta_t)}_{i}, i=1,\ldots,d$ is given by\\
  \STATE {\footnotesize 
  \begin{align*}
   &\left[\left(\mEt^i - \oEt^i\right)^2+\epsilon\right]^{-1}\left(\mEt^i - \oEt^i\right) && (U_1^*.G_1)\\   
   &\texttt{arctan}\left( \frac{\mEt^i - \oEt^i}{\sqrt{\epsilon}} \right) && (U_2^*.G_1)\\
   &\texttt{erf}\left(\alpha (\mEt^i - \oEt^i)\right) && (U_2^*.G_2)\\
   &\texttt{gd}\left(\beta (\mEt^i - \oEt^i)\right) && (U_2^*.G_3)
  \end{align*}}
  \STATE $z_{t+1} = (1-C\lambda_t)z_t$
  \STATE 
  \IF{no more improvement on training set}
  \STATE exit
  \ENDIF
  \ENDFOR
  \ENDFOR
  \end{algorithmic}
\caption{\label{fig:alg}Summary of the proposed algorithms.}
\end{figure}

\subsection{Computational Complexity}
The proposed algorithms simply transforms the gradient of the log-likelihood, and each transformation function can be computed in constant time, therefore all of them have the same time complexity as SGD ($O(d)$ per update). In cases where only indicator features are defined, the training process can be accelerated by pre-computing the values of $\grad^2G$ or $\grad G$ for variables within range $\left[-1, 1\right]$ and keep them in a table. During the actual training, the transformed gradient values can be found simply by looking up the table after rounding-off to the nearest entry. This way we do not even need to compute the function values on the fly, which significantly reduces the computational cost.

\section{Experiments}\label{sec:experiments}
\subsection{Settings}
We conduct our experiments based on the following settings:

\emph{Implementation}: We implemented our algorithms based on the CRF suite toolkit~\cite{Okazaki:2007}, in which SGD for the log-likelihood loss with L2 regularization is already implemented. This can be easily done since our algorithm only requires to modify the original gradients. Other parts of the code remain unchanged.

\emph{Task}: Our experiments are conducted for the widely used CoNLL 2000 chunking shared task~\cite{Sang:2000}. The training and test data contain 8939 and 2012 sentences respectively. For fair comparison, we ran our experiments with two standard linear-chain CRF feature sets implemented in the CRF suite. The smaller feature set contains 452,755 features, whereas the larger one contains 7,385,312 features.

\emph{Baseline algorithms}: We compare the performance of our algorithms summarized in Figure~\ref{fig:alg} with SGD, L-BFGS, and the Passive-Aggressive (PA) algorithm. Except for L-BFGS which is a second-order batch-learning algorithm, we randomly shuffled the data and repeated experiments five times.

\emph{Hyper-parameter selection}: For convex function $G_1$ we choose $\epsilon=0.1$, correspondingly the \texttt{arctan} function in $U_2^*.G_1$ is $\arctan(3.16u)$. We have also experimented the function $\arctan(10u)$ for this update, following the heuristic that the transformation function $\frac{u}{u^2+0.1}$ given by $\grad^2 G_1$ and $\arctan(10u)$ have consistent gradients at zero, since the $U_2^*$ update imitates the behavior of $U_1^*$ (the two choices  of the \texttt{arctan} functions are denoted by \texttt{arctan.1} and \texttt{arctan.2} respectively). Following the same heuristic, we choose $\alpha=5\sqrt{\pi}\approx8.86$ for the \texttt{erf} function and $\beta=10$ for the \texttt{gd} function.

\subsection{Results}
\begin{figure}[ht]
\centering
%\makebox[\textwidth] {
  \begin{subfigure}[b]{0.3\textwidth}
  \includegraphics[width=\textwidth]{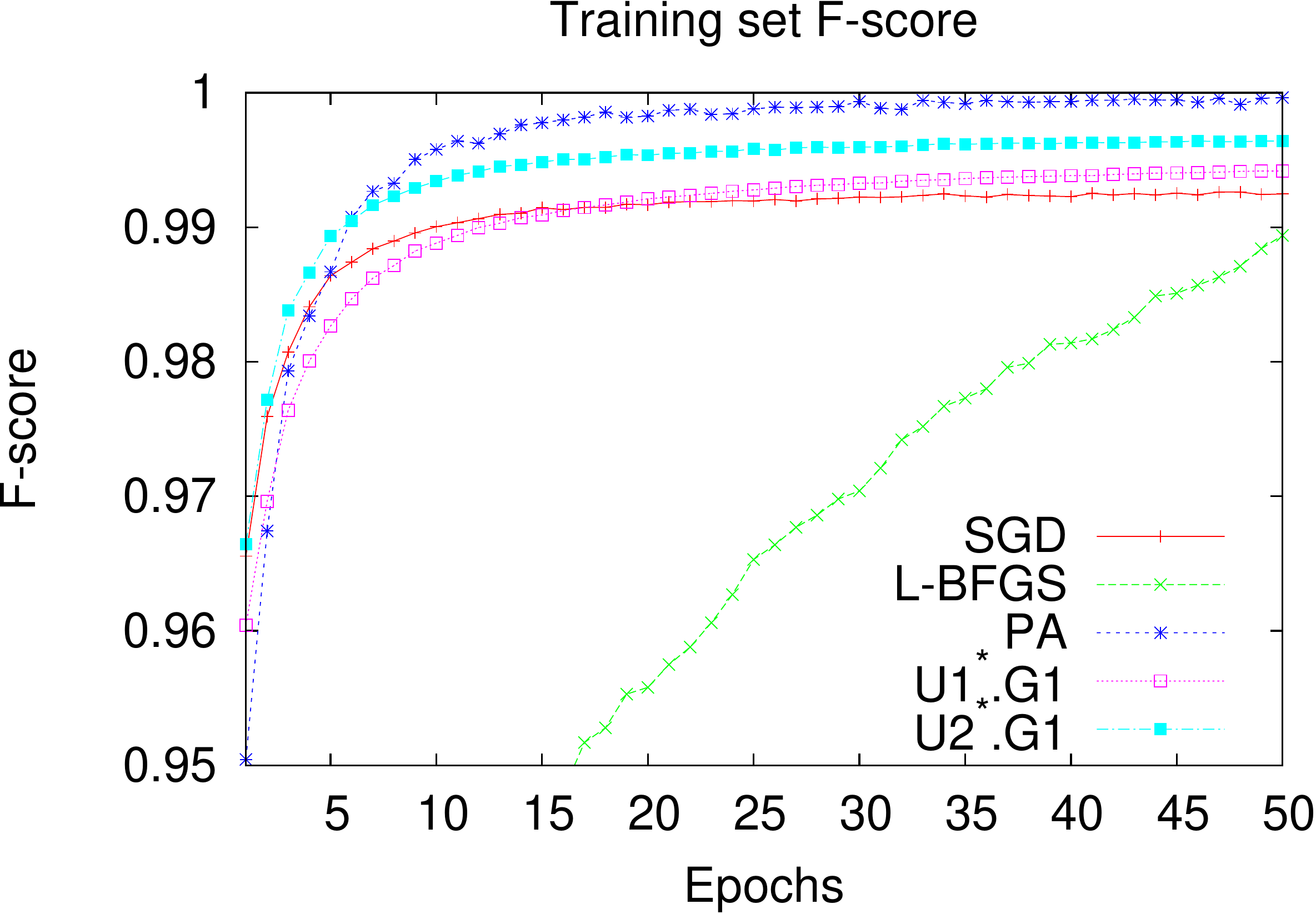}
  %\caption{Training set}
  \end{subfigure} 
  \begin{subfigure}[b]{0.3\textwidth}
  \includegraphics[width=\textwidth]{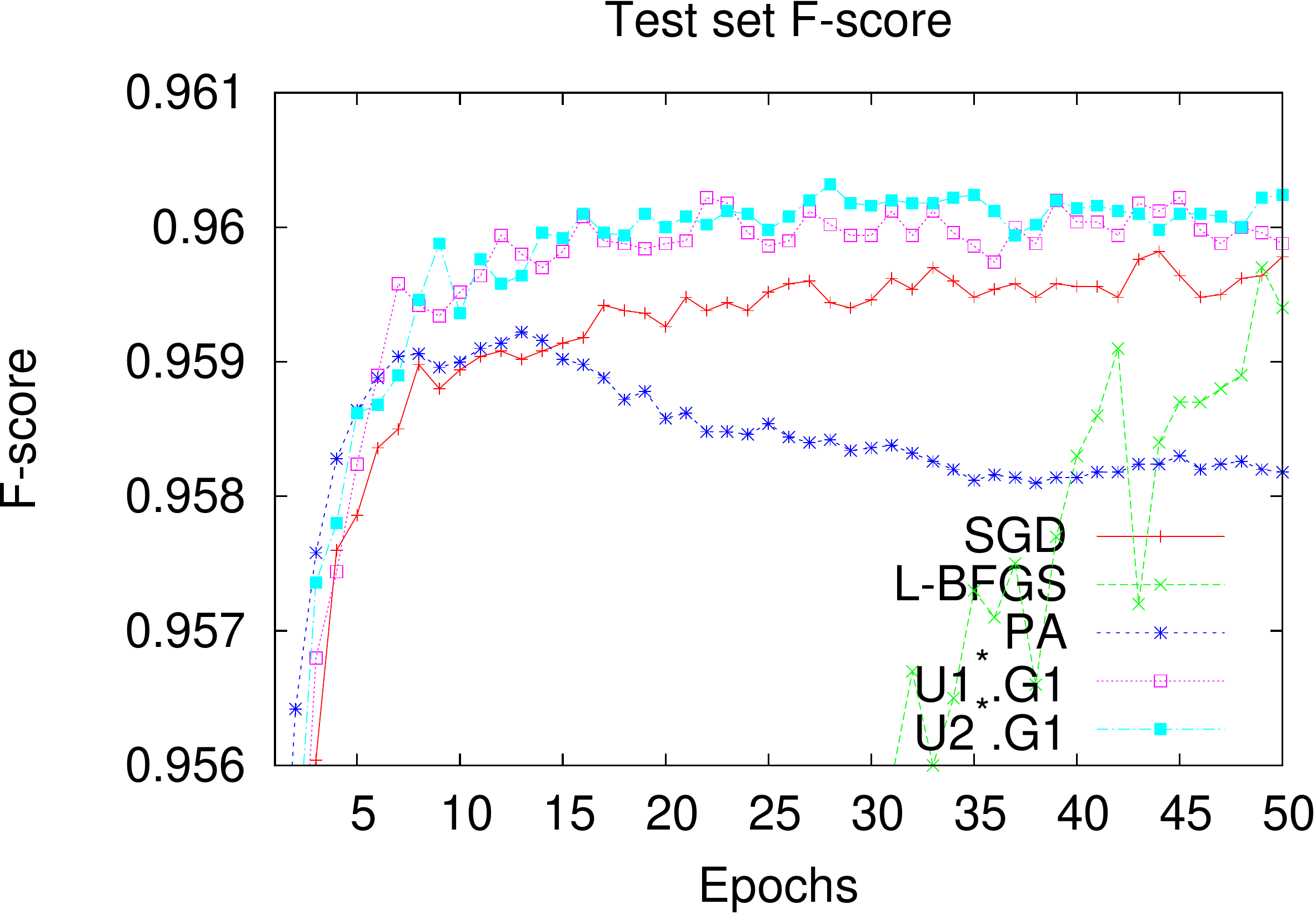}
  %\caption{Test set}
  \end{subfigure} %}
  \caption{F-scores of training and test sets given by the baseline and proposed algorithms, using the small feature set.} \label{fig:base_small}
\end{figure}

\begin{figure}[ht]
\centering
%\makebox[\textwidth] {
  \begin{subfigure}[b]{0.3\textwidth}
  \includegraphics[width=\textwidth]{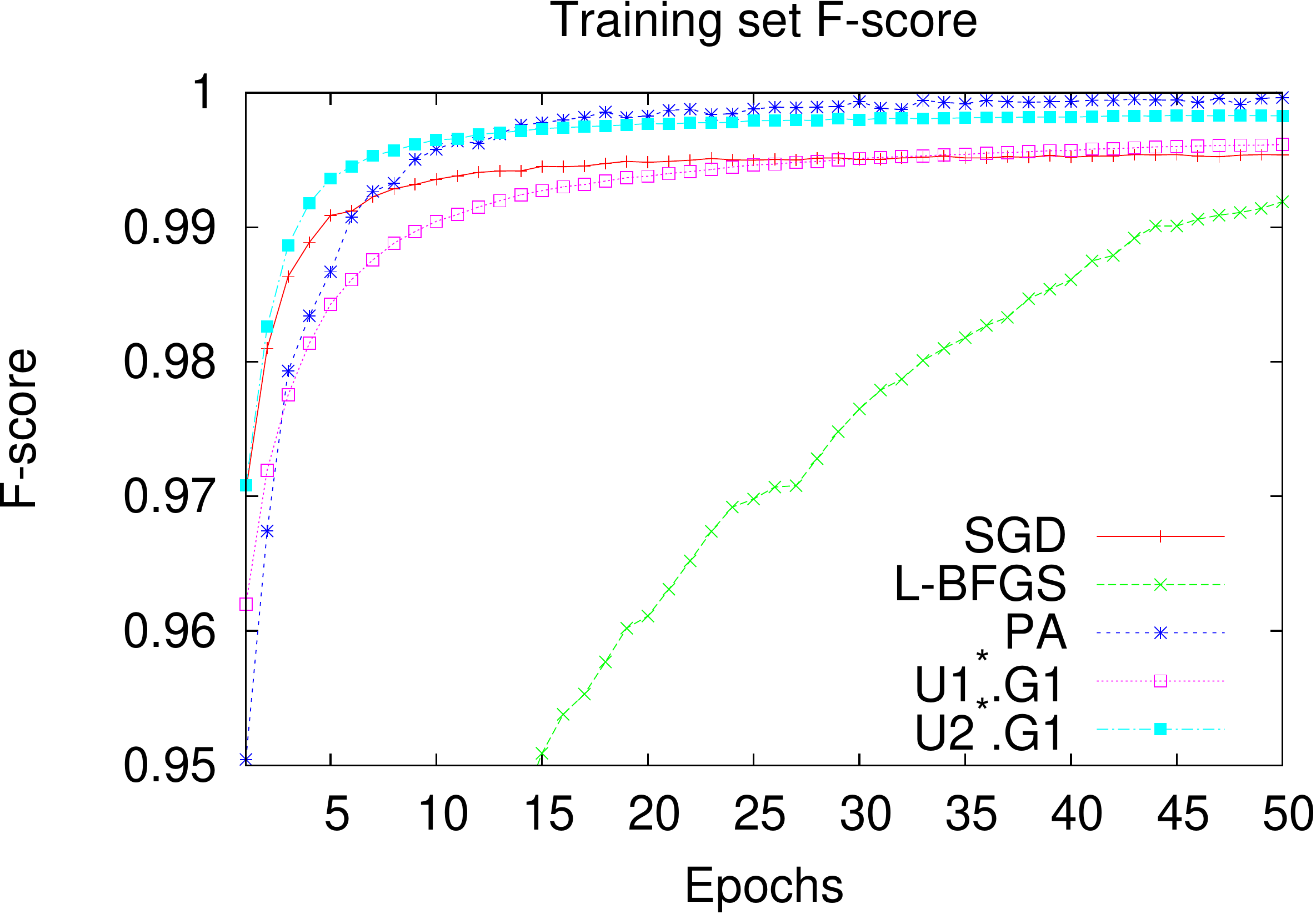}
  %\caption{Training set} \label{fig:tune_curve}
  \end{subfigure} 
  \begin{subfigure}[b]{0.3\textwidth}
  \includegraphics[width=\textwidth]{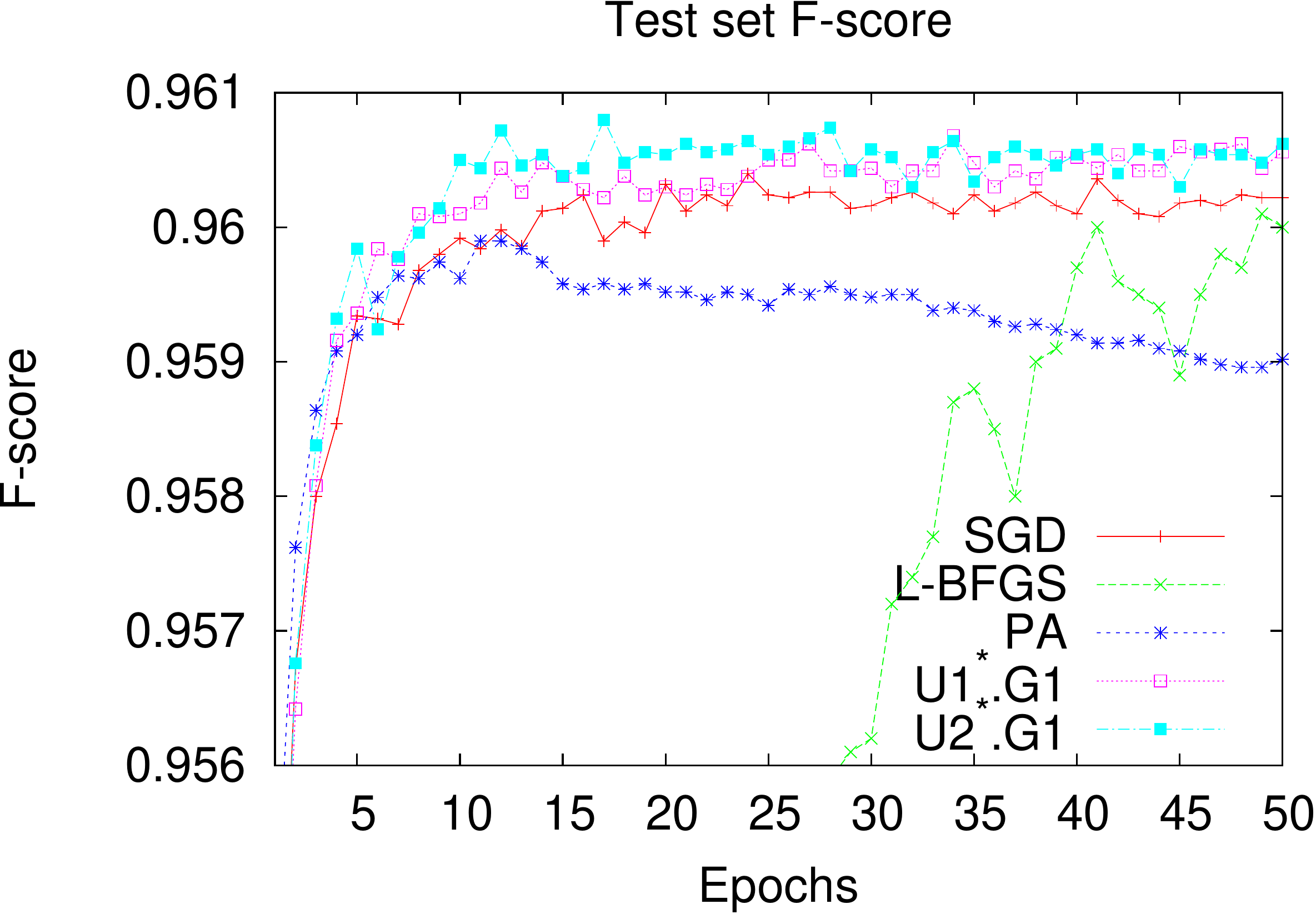}
  %\caption{Test set} \label{fig:held_curve}
  \end{subfigure} %}
  \caption{F-scores of training and test sets given by the baseline and proposed algorithms, using the large feature set.} \label{fig:base_large}
\end{figure}

\emph{Comparison with the baseline}: We compare the performance of the proposed and baseline algorithms on the training and test sets in Figure~\ref{fig:base_small} and Figure~\ref{fig:base_large}, corresponding to small and large feature sets respectively. The plots show the F-scores on the training and test sets for the first 50 epochs. To keep the plots neat, we only show the average F-scores of repeated experiments after each epoch, and omit the standard deviation error bars. For $U_2^*.G_1$ update, only $\arctan.1$ function is reported here. From the figure we observe that:

1. The strongest baseline is given by SGD. By comparison, PA appears to overfit the training data, while L-BFGS converges very slowly, although eventually it catches up.

2. Although the SGD baseline is already very strong (especially with the large feature set), both the proposed algorithm $U_1^*.G_1$ and $U_2^*.G_1$ outperform SGD and stay on top of the SGD curves most of the time. On the other hand, the $U_2^*.G_1$ update appears to be a little more advantageous than $U_1^*.G_1$.\\

\begin{figure}[ht]
\centering
%\makebox[\textwidth] {
  \begin{subfigure}[b]{0.3\textwidth}
  \includegraphics[width=\textwidth]{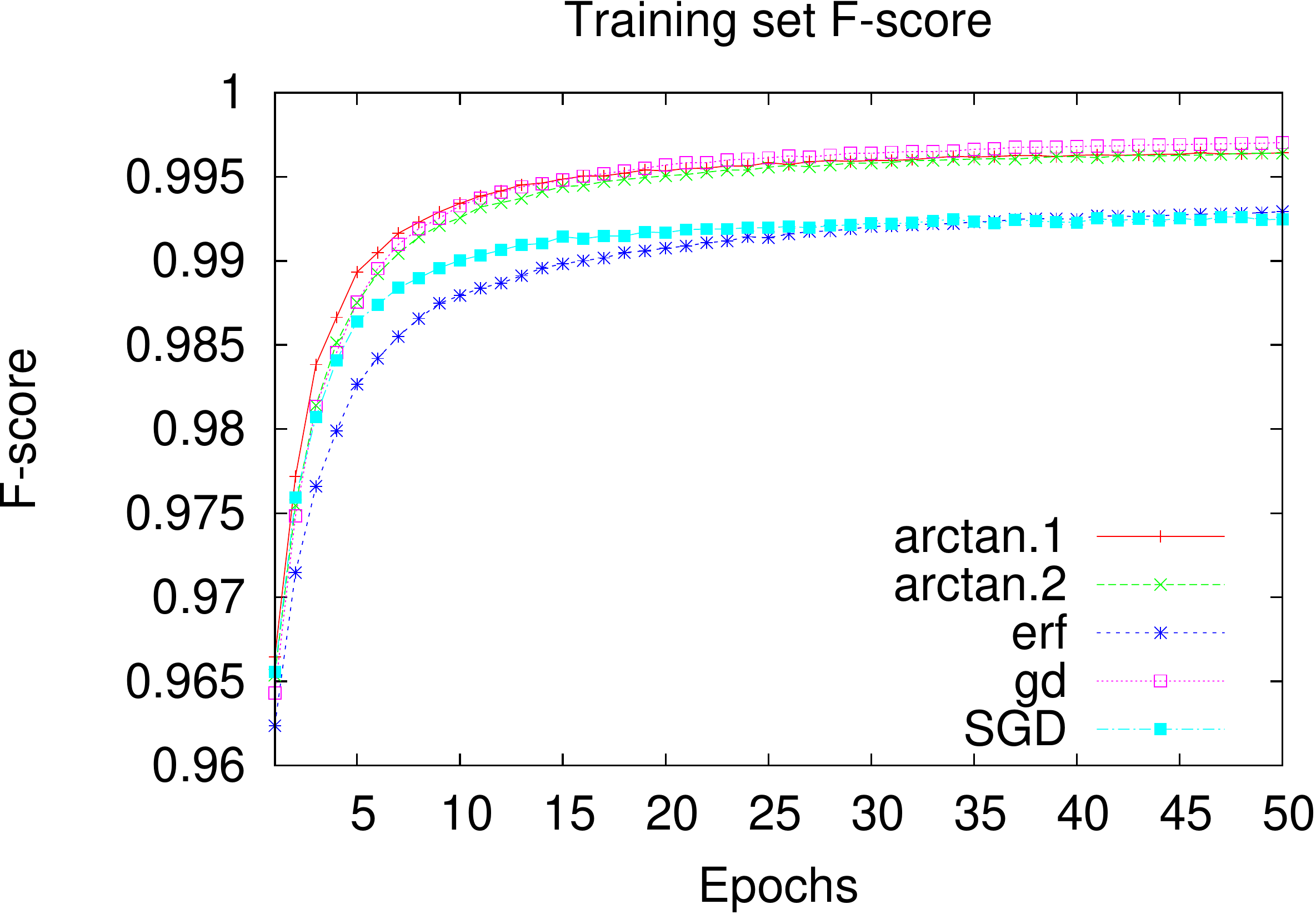}
  %\caption{Training set} \label{fig:tune_curve}
  \end{subfigure} 
  \begin{subfigure}[b]{0.3\textwidth}
  \includegraphics[width=\textwidth]{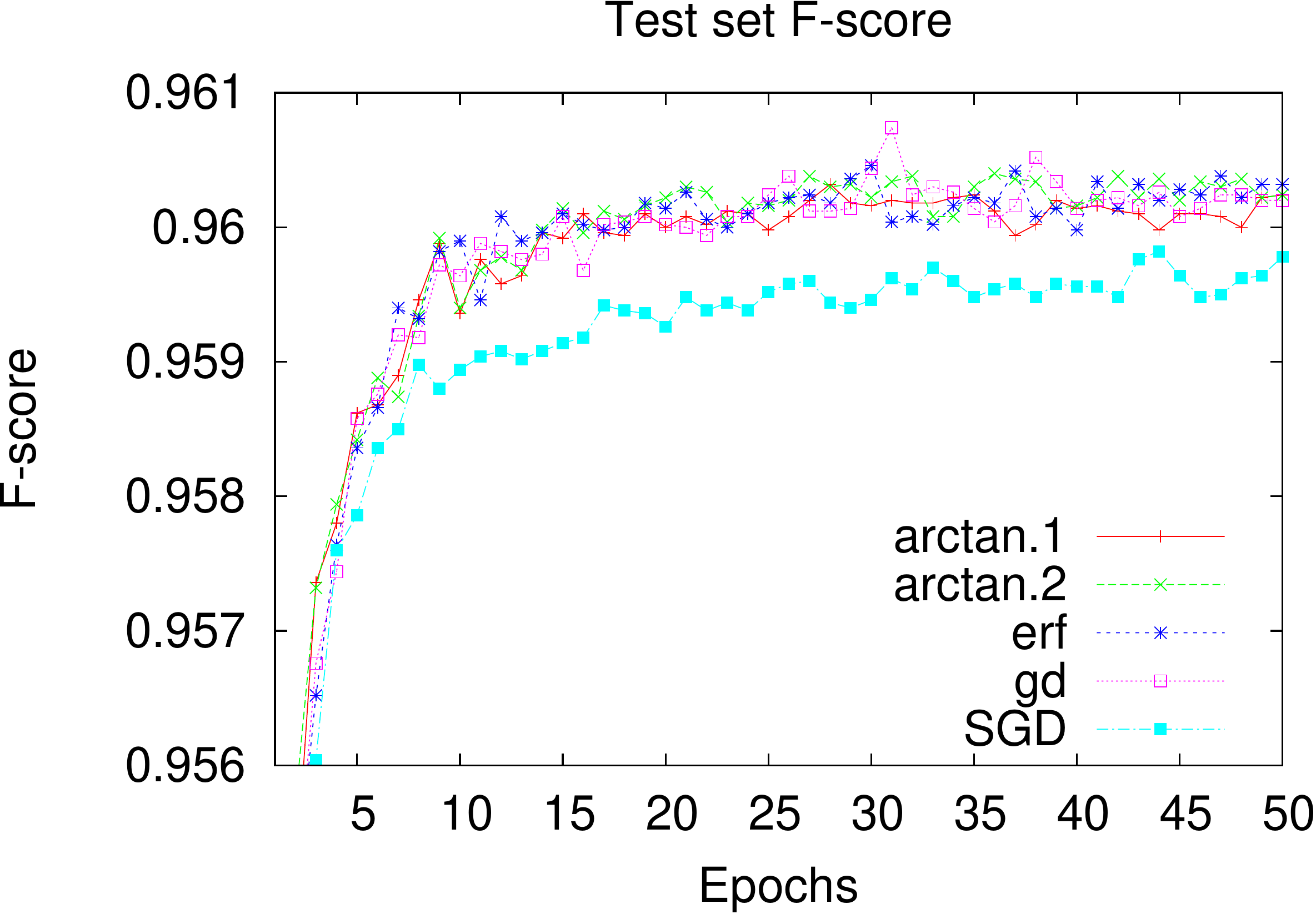}
  %\caption{Test set} \label{fig:held_curve}
  \end{subfigure} %}
  \caption{F-scores of training and test sets given by $U_2^*.G1$, $U_2^*.G1$ and $U_2^*.G3$, using the small feature set.} \label{fig:sig_small}
\end{figure}

\begin{figure}[H]
\centering
%\makebox[\textwidth] {
  \begin{subfigure}[b]{0.3\textwidth}
  \includegraphics[width=\textwidth]{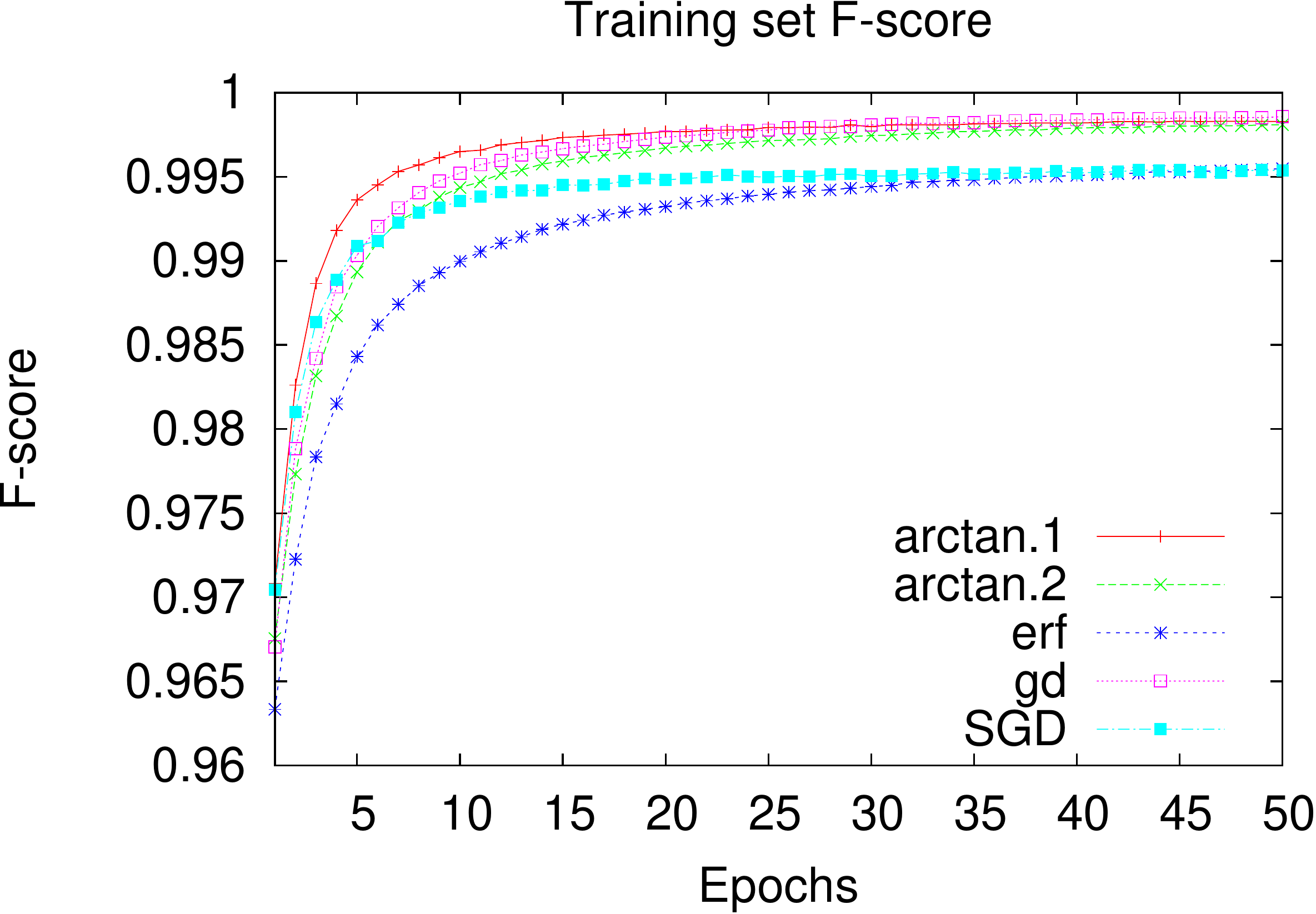}
  %\caption{Training set} \label{fig:tune_curve}
  \end{subfigure} 
  \begin{subfigure}[b]{0.3\textwidth}
  \includegraphics[width=\textwidth]{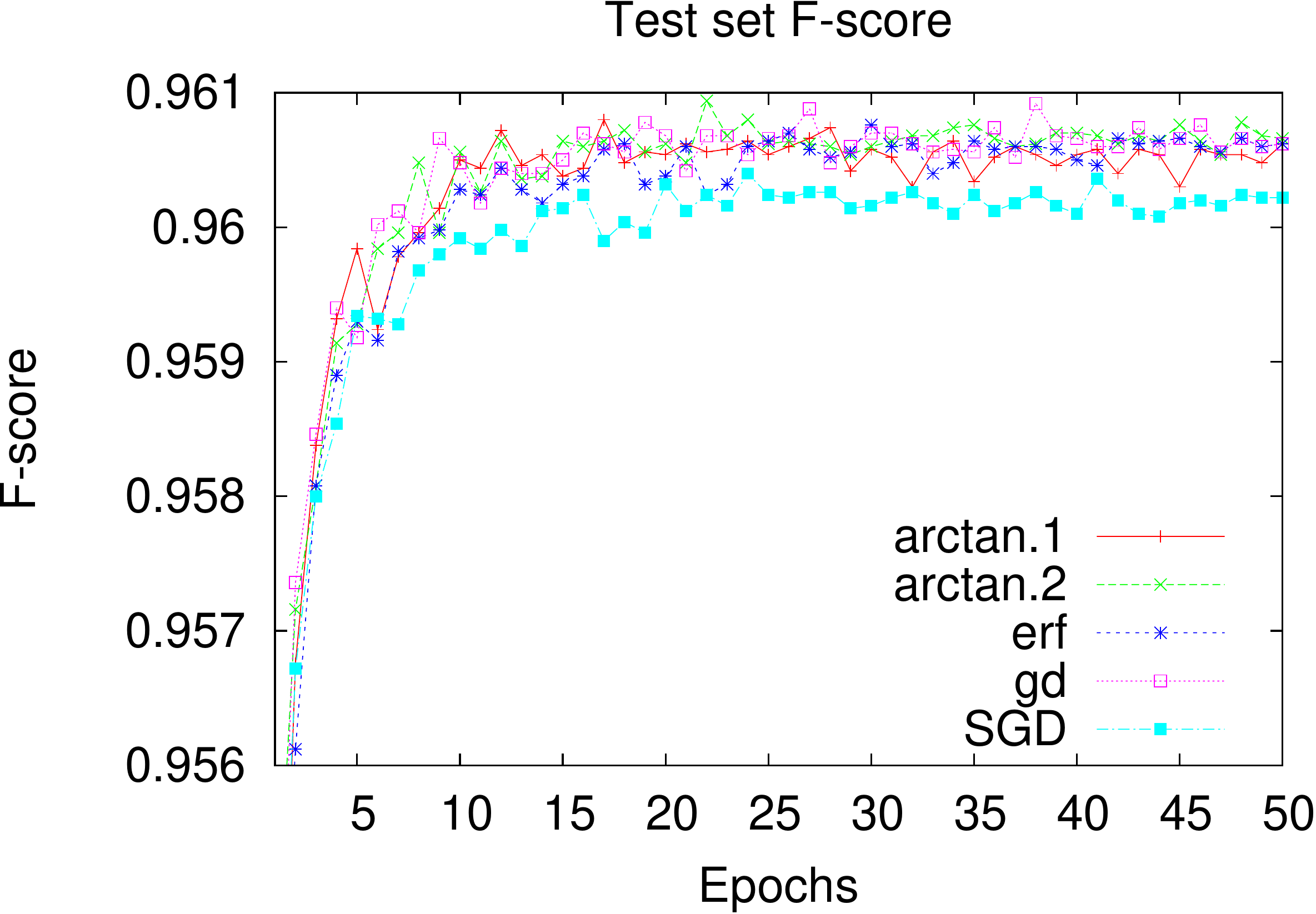}
  %\caption{Test set} \label{fig:held_curve}
  \end{subfigure} %}
  \caption{F-scores of training and test sets given by $U_2^*.G1$, $U_2^*.G1$ and $U_2^*.G3$, using the large feature set.} \label{fig:sig_large}
\end{figure}

\noindent\emph{Comparison of the sigmoid functions}: Since \texttt{arctan}, \texttt{erf} and \texttt{gd} are all sigmoid functions and it is interesting to see how their behaviors differ, we compare the updates $U_2^*.G1$ (for both $\arctan.1$ and $\arctan.2$), $U_2^*.G_2$ and $U_2^*.G_3$ in Figure~\ref{fig:sig_small} and Figure~\ref{fig:sig_large}.  The strongest baseline, SGD, is also included for comparison. From the figures we have the following observations:

1. As expected, the sigmoid functions demonstrated similar behaviors, and performances of their corresponding updates are almost indistinguishable. $U_2^*.G_2$. Similar to $U_2^*.G1$, $U_2^*.G2$ and $U_2^*.G3$ both outperformed the SGD baseline.

2. Performances of $U_2^*.G1$ given by \texttt{arctan} functions are insensitive to the choice of the hyper-parameters. Although we did not run similar experiments for \texttt{erf} and \texttt{gd} functions, similar properties can be expected from their corresponding updates.

Finally, we report in Table~\ref{table:fscore} the F-scores on the test set given by all algorithms after they converge.
\begin{table}[ht]
{\small
  \begin{tabular}{ c c c }
  Algorithm & F-score \% (small) & F-score \% (large)\\
  \hline
  SGD  & 95.98$\pm$0.02 & 96.02$\pm$0.01\\
  PA & 95.82$\pm$0.04 & 95.90$\pm$0.03 \\
  L-BFGS & 96.00 & 96.01\\
  $U_1^*.G_1$ & 95.99$\pm$0.02 & 96.06$\pm$0.03\\
  $U_2^*.G_1$ & 96.02$\pm$0.02 & 96.06$\pm$0.02\\
  $U_2^*.G_1'$ & 96.03 $\pm$ 0.01 & 96.06$\pm$0.03 \\
  $U_2^*.G_2$ & 96.03$\pm$0.02 & 96.06$\pm$0.02\\
  $U_2^*.G_3$ & 96.02$\pm$0.02 & 96.06$\pm$0.02\\
  \hline  
  \end{tabular}}
  \caption{F-scores on the test set after algorithm converges, using the small and large feature sets. $U_2^*.G_1$ is the update given by \texttt{arctan.1}, and $U_2^*.G_2$ given by \texttt{arctan.2}.}\label{table:fscore}
\end{table}

\section{Conclusion}
We have proposed a novel parameter estimation framework for CRF. By defining loss functions using the Bregman divergences, we are given the opportunity to select convex functions that transform the gradient of the log-likelihood loss, which leads to more effective parameter learning if the function is properly chosen. Minimization of the Bregman loss function is made possible by NGD, thanks to the structure of exponential families. We developed several parameter update strategies which approximately incorporates the second-order information of the log-likelihood, and outperformed baseline algorithms that are already very strong on a popular text chunking task.

Proper choice of the convex functions is critical to the performance of the proposed algorithms, and is an interesting problem that merits further investigation. While we selected the convex functions with the motivation to reduce the types of updates and incorporate approximate second-order information, there are certainly more possible choices and the performance could be improved via careful theoretical analysis. On the other hand, instead of choosing a convex function apriori, we may rely on some heuristics from the actual data and choose a function tailored for the task at hand.

\end{document}